\documentclass[letterpaper, 10 pt, journal, twoside]{IEEEtran}

\usepackage{xcolor,soul,framed} 
\colorlet{shadecolor}{yellow}
\usepackage[pdftex]{graphicx}
\graphicspath{{../pdf/}{../jpeg/}}
\DeclareGraphicsExtensions{.pdf,.jpeg,.png}

\usepackage[cmex10]{amsmath}
\usepackage{array}
\usepackage{booktabs} 
\usepackage{siunitx} 
\usepackage{makecell} 
\usepackage{pbox} 
\usepackage{threeparttable} 
\usepackage{multirow}
\usepackage{mdwmath}
\usepackage{mdwtab}
\usepackage{eqparbox}
\usepackage{url}
\usepackage{algorithm}
\usepackage{algpseudocodex}
\usepackage[colorlinks=true, linkcolor=blue, citecolor=blue, urlcolor=blue, hidelinks]{hyperref}
\usepackage{amssymb}
\ifCLASSOPTIONcompsoc
\usepackage[caption=false, font=normalsize, labelfont=sf, textfont=sf]{subfig}
\else
\usepackage[caption=false, font=footnotesize]{subfig}
\fi

\begin{document}
    \title{AELF: Active Terrain-Aware UAV Exploration Using a Lightweight Framework}
  \author{
  Mingyang Li, Xiao Cao, Yi Luo, Yuting Tao, Hongjie Huang, Bowen Xu, Peng Lu

  \thanks{This work was supported by General Research Fund under Grant 17204222, and in part by the Seed Fund for Collaborative Research and General Funding Scheme-HKU-TCL Joint Research Center for Artificial Intelligence. \it{(Mingyang Li and Xiao Cao contributed equally to this work.)(Corresponding author: Peng Lu.)} The authors are with the Adaptive Robotic Controls Lab (ArcLab), Department of Mechanical Engineering, The University of Hong Kong, Hong Kong, SAR 999077, China (e-mail: myli0823@connect.hku.hk; u3009678@connect.hku.hk; u3010355@connect.hku.hk; taoyt@connect.hku.hk; hjhuang@connect.hku.hk; link.bowenxu@connect.hku.hk; lupeng@hku.hk)}} 


\maketitle
\pagestyle{empty}
\thispagestyle{empty}

\begin{abstract} 
This paper presents an efficient and robust approach for autonomous aerial exploration in complex environments. Instead of relying on conventional 3D voxel-based maps for exploration, we utilize a lightweight 2.5D map framework that reduces computational overhead while preserving essential topographic information. More importantly, this representation systematically eliminates the blind voxel problem commonly encountered in voxel-based exploration due to occlusions and limited visibility. Unlike previous methods that rely on a fixed altitude scale for exploration, our approach requires no prior height information and actively adapts to terrain elevation changes during exploration. To further enhance exploration efficiency while preserving the generality of 3D exploration, we perform 2D path planning on the 2.5D map framework, which is then lifted to 3D to evaluate the cost between viewpoints by taking into account terrain traversability, thereby reducing frequent altitude changes and enabling smoother exploration. Extensive experiments conducted in both simulated and real-world environments demonstrate the effectiveness and advantages of our approach over existing solutions. Our approach demonstrates fast exploration of over 2000~$\text{m}^2$ of cluttered forest and sloped mountainous terrain with elevation changes of up to 10~m, reaching a maximum speed of 5~m/s in real-world tests.
\end{abstract}


\IEEEpeerreviewmaketitle

\section{Introduction}
\IEEEPARstart{I}{n} real-world applications such as search and rescue, environmental monitoring, infrastructure inspection, and space exploration, autonomous exploration plays a critical role. These tasks often involve complex terrains—flat areas, slopes, and rugged regions—where robots equipped with advanced sensors can map environments and detect obstacles in real time. In time-sensitive scenarios like disaster response, precise geometric reconstruction is often unnecessary; what matters more is the rapid acquisition of traversable 3D structures to support fast decision-making. Many of these missions rely on multi-robot collaboration, where aerial robots provide quick scene coverage despite limited endurance. Efficient use of their capabilities offers valuable environmental priors for planning and deploying ground-based systems.

Several existing methods focus on autonomous exploration in 2D environments and have achieved promising results in flat or structured terrains \cite{yamauchi1997frontier, ExplorDRL2019, li2025daptp, hu2020voronoi}. However, they perform poorly in environments with unstructured topography, such as hills or sloped terrain. In such cases, 2D exploration techniques fail to capture vertical structure, often resulting in incomplete map coverage and inefficient navigation. Moreover, relying solely on 2D sensors limits the robot’s ability to accurately distinguish between traversable and non-traversable areas in three-dimensional space, potentially compromising safety in real-world deployments \cite{cao2024hma}. These limitations reduce overall exploration effectiveness and hinder reliable operation in rugged or uneven environments, highlighting the need for more advanced strategies capable of handling 3D spatial complexity.

To address this, modern systems increasingly use 3D sensors such as LiDARs and RGB-D cameras \cite{cao2021tare, zhou2024oto, lu2024fapp}. These sensors offer richer spatial information than 2D ones: LiDAR produces accurate geometric point clouds, while RGB-D cameras offer both geometric and texture cues, enabling more informed decision-making in visually complex spaces. Most existing 3D exploration methods model the environment using occupancy grid maps with predefined resolutions and height limits. The aim is to estimate the occupancy state of each grid cell during navigation, and the exploration size is preset to ensure that the unknown occupancy grid fully encloses the target regions. However, without prior knowledge of the environment’s geometry, such preset height constraints can lead to high memory usage or missed regions, reducing exploration accuracy and completeness.

\begin{figure}[t]
    \centering
    \includegraphics[width=\columnwidth]{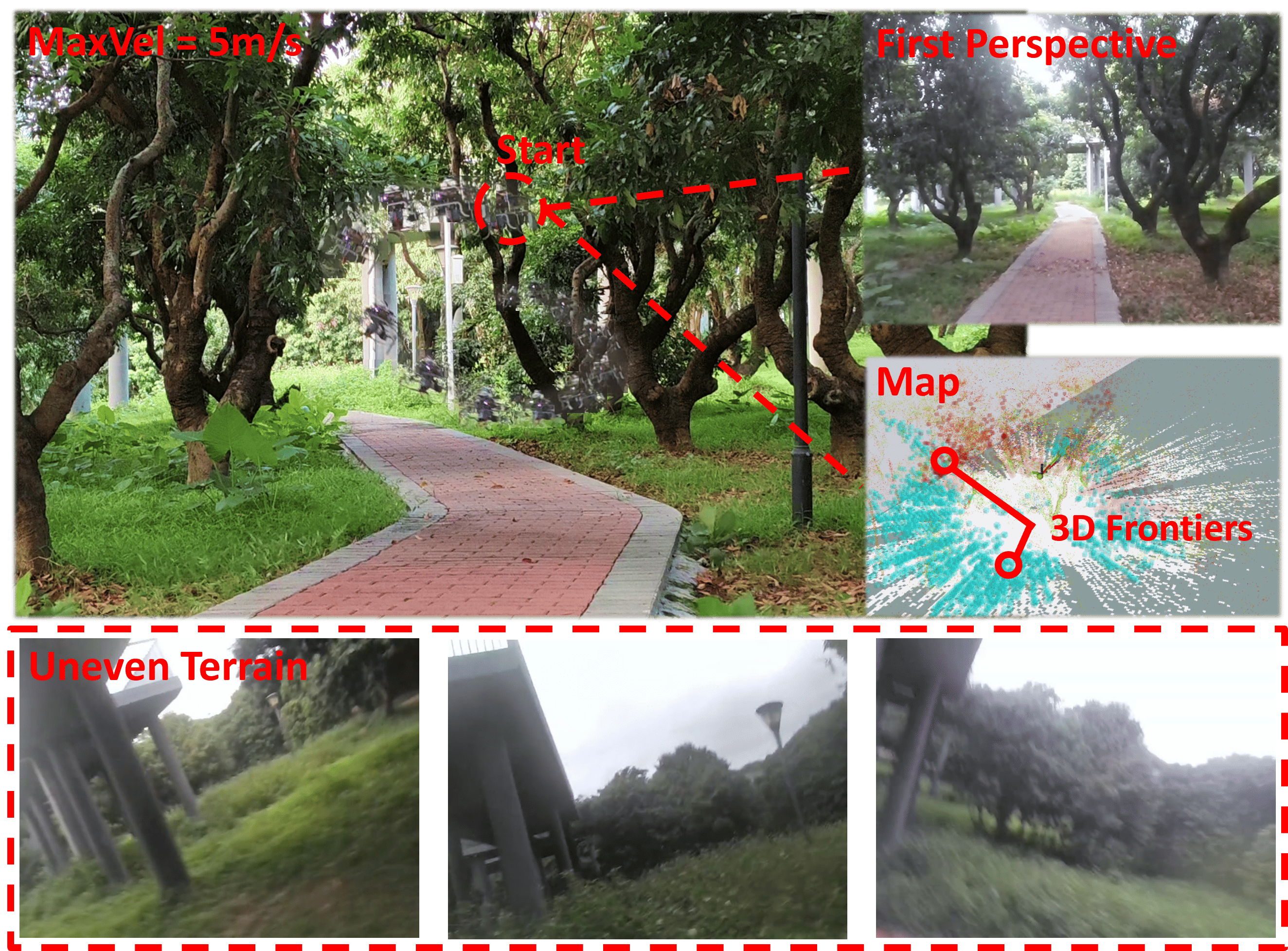}
    \caption{A UAV autonomously explores an unknown forest environment characterized by uneven terrain and complex obstacles.}
    \label{fig: cover}
\end{figure}

In 3D exploration, some methods improve efficiency by discarding clusters with few frontiers or those deemed unreachable \cite{zhou2021fuel, zhou2023racer, luperto2024estimating}. This allows the robot to focus on more informative and accessible areas, enhancing planning efficiency. In contrast, other approaches \cite{zhou2024oto, bartolomei2023fast} do not ignore smaller or marginal unexplored regions. Instead, they incorporate these areas into the planning process, aiming to clear space more systematically rather than exclusively prioritizing regions with the highest expected information gain. While such methods are applicable to 3D environments, they are typically designed for scenarios with relatively structured spatial layouts \cite{zhao2023autonomous}. In environments with complex and unstructured obstacles, such as forests, cluster reachability is often assessed during execution rather than prior to planning. This reactive checking can lead to inefficiencies, such as wasted planning cycles or idle waiting for unreachable clusters to be cleared, thereby disrupting exploration continuity.

Although 2D maps offer substantial advantages over 3D voxel grids in terms of reduced memory usage and computational load, effective conversion from 3D to 2D remains challenging. The most common approach projects the 3D voxel map, such as OctoMap \cite{hornung2013octomap}, onto a 2D plane. A more advanced method, HMAPs \cite{yang2018efficient}, supports conversion from 3D to both 2.5D and 2D maps via downward projection. However, these projection-based techniques have inherent limitations. For example, any obstacle on the ground or ceiling is also projected onto the 2D map, potentially causing ambiguity in occupancy interpretation. Additionally, any structure outside the projection range is omitted entirely, causing incomplete environment representation. These drawbacks limit the generalization of such methods in environments with rich vertical structures or overhanging elements.
 
To address the limitations mentioned above, especially in complex environments with significant terrain variation and unknown scale, we propose \textbf{AELF} (\textbf{A}ctive \textbf{E}xploration using a \textbf{L}ightweight \textbf{F}ramework). In this framework, 3D frontiers are extracted from a 2D grid map, and their corresponding elevations are retrieved from a height map. Regions that lack sufficient traversable space are filtered out using a predefined safety threshold to ensure safe and feasible planning. By utilizing spatial information, we estimate local terrain slopes to evaluate the terrain traversability. This enables the integration of terrain-aware cost terms into the planning process, which encourages smoother and more consistent exploration trajectories. Our main contributions are as follows:
\begin{itemize}
    \item A projection-free method is used to convert 3D maps into 2D representations by extracting free space directly, enabling robust exploration in complex and unstructured outdoor environments. Terrain elevation is adaptively corrected using a sensor-aware model to prevent height bootstrapping during exploration.
    \item A 2.5D map framework is used for frontier detection and 3D path generation to accelerate exploration planning. This design systematically eliminates the blind voxel problem commonly encountered in outdoor exploration.
    \item By considering traversability information, the system retains 3D exploration generality while using 2D terrain cues for smoother motion and fewer altitude changes.
    \item Extensive simulations and real-world experiments validate the framework, with the system running on a self-built UAV and tested in challenging forest exploration at speeds up to 5 m/s, as shown in Fig. \ref{fig: cover}.
\end{itemize}

\section{Related work}
\subsection{Map Representation for Autonomous Systems}
Maps play a fundamental role in autonomous systems such as UAVs and ground vehicles, enabling key functionalities like localization, motion planning, and obstacle avoidance. Among various map representations, 2D and 3D metric grid maps are the most widely adopted due to their simplicity and compatibility with common algorithms. While 3D maps offer detailed spatial representation, 2D maps are more efficient in terms of computation and memory. Combining both is often advantageous, but maintaining separate 2D and 3D mapping pipelines, such as by projecting 3D data onto a 2D plane or using different sensors to generate separate maps, increases system complexity and may lead to inconsistencies that degrade planning accuracy and robustness \cite{Fred2024voxel}.

To address this, unified mapping frameworks for autonomous platforms aim to generate maps directly from raw sensor data. Prior studies have converted point clouds into 2D maps \cite{tao2024mobile, chen2024lidar} or extracted elevation values as 2D height images \cite{wang2025terrain}. However, these often struggle with vertical structures such as overhangs and ceilings, leading to incomplete or distorted maps. Projection-based methods, such as filtering out floor points before projection \cite{li2023advanced}, are usually sensor-specific and less generalizable across different autonomous systems.

Translating 3D voxel maps into 2D representations remains relatively underexplored, especially for UAVs operating in large-scale, unstructured environments. A common method is to project a fixed-height slice onto a 2D plane, as in OctoMap \cite{hornung2013octomap}. SkiMap \cite{de2017skimap} improves efficiency by classifying voxels and updating a specific SkipList Tree layer. However, such projections inherently discard vertical information, misrepresenting environments with elevation variation. They may misinterpret ground or ceiling voxels and fail to detect obstacles beyond a fixed height range. While such limitations are often acceptable in flat indoor environments, they can pose significant challenges in large-scale outdoor scenarios.

\begin{figure*}[t]
    \centering
    \includegraphics[width=\textwidth]{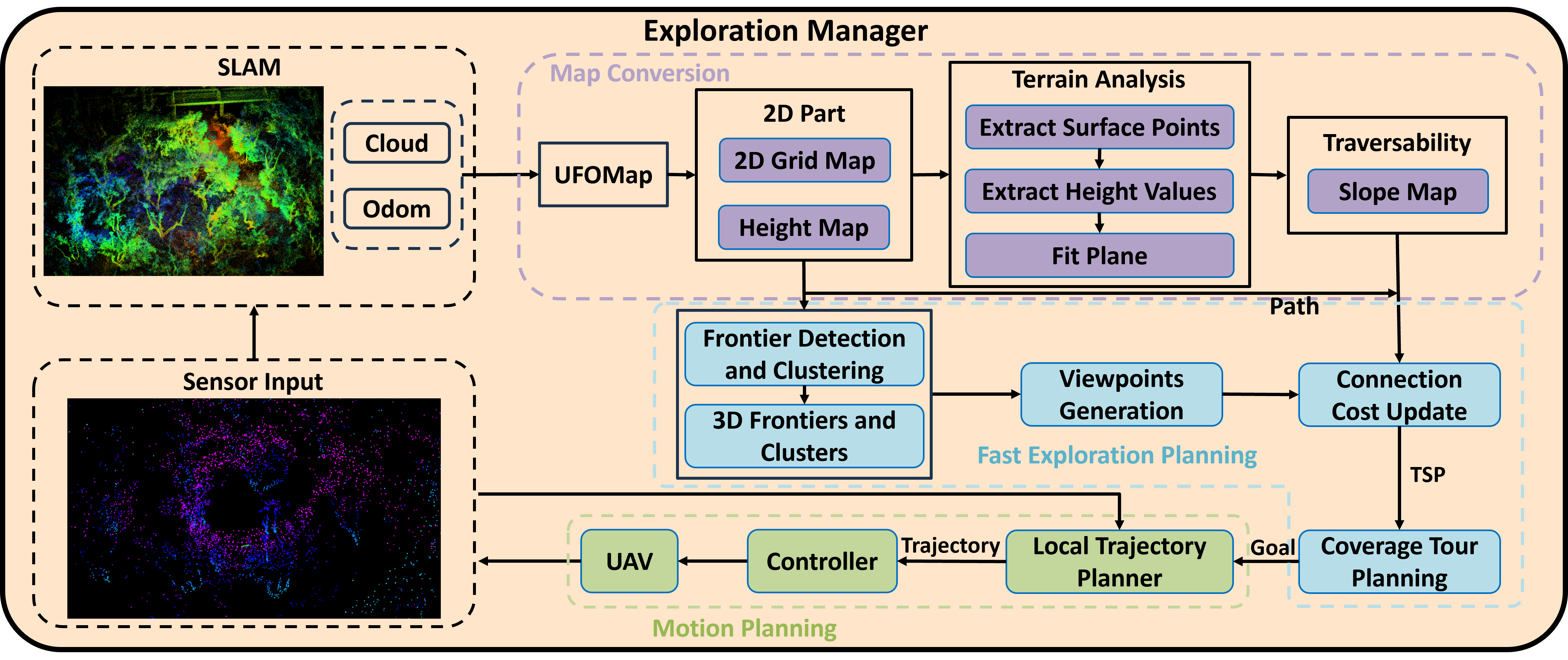}
    \caption{System architecture of the proposed 2.5D terrain-aware exploration framework. The system consists of three main modules: SLAM, Exploration Planner (Map Conversion and Fast Exploration Planning), and Motion Planning.}
    \label{fig: framework}
\end{figure*}

To address this, we utilize a lightweight, task-oriented mapping framework for autonomous exploration. Our method merges the efficiency of 2D frontier detection with the 3D spatial fidelity, enabling effective exploration in environments with complex elevation changes.

\subsection{Sampling-based Exploration for Autonomous Systems}
Sampling-based exploration methods are widely adopted in autonomous systems, particularly for UAVs operating in unknown or large-scale environments \cite{bircher2016receding, dang2020graph, xu2021autonomous, border2024osprey}. These approaches select candidate viewpoints within the robot’s known free space and rank them using spatial utility or cost functions, guiding the robot to sequentially explore high-priority areas. A representative method is RH-NBVP \cite{bircher2016receding}, which uses Rapidly-exploring Random Trees (RRT) \cite{lavalle1998rapidly} to identify the next-best view. After each movement toward the most informative node, the RRT is reconstructed from the robot’s current position using a receding horizon strategy. While offering probabilistic completeness, this approach can lead to redundant paths and high computational load, which are problematic for real-time autonomous UAV systems.

To improve exploration efficiency, many extensions have been developed for autonomous aerial platforms. Witting et al. \cite{witting2018history} reuse previously visited viewpoints as RRT seed nodes, which accelerates redirection toward informative regions when the robot reaches dead ends. The AEP framework \cite{selin2019efficient} integrates frontier-based global planning with local NBVP-style strategies to avoid local entrapment. GBP \cite{dang2020graph} addresses the next-best-view problem using Rapidly-Random Graphs (RRG) \cite{karaman2011sampling}, constructing a dense RRG for local planning that ensures high information gain and collision avoidance, while a sparse RRG is used for global guidance. However, the dual-RRG structure incurs a high computational cost due to evaluating information gain for all candidate nodes. To mitigate this, DSVP \cite{zhu2021dsvp} proposes a biased sampling approach that guides RRT growth using both local and global frontiers, avoiding per-node evaluation while retaining exploration performance.

\subsection{Frontier-based Exploration for Autonomous Systems}
Another widely used strategy for autonomous UAV exploration is frontier-based exploration \cite{zhou2021fuel, cao2021exploring, yu2023echo}, which guides robots to the boundaries between known and unknown space. First introduced by Yamauchi \cite{yamauchi1997frontier}, the method selects the nearest frontier as the next goal. Gao et al. \cite{gao2018improved} improve it by adding heading and rotation costs for smoother motion, while Freda et al. \cite{freda2005frontier} use Sensor-based Random Trees (SRT) to bias exploration toward frontiers. While most traditional methods operate in 2D, researchers have extended them to 3D. Shen et al. \cite{shen2012stochastic, shen2012autonomous} apply stochastic differential equations to detect areas with particle expansion, improving target selection under sparse data. Dornhege et al. \cite{dornhege2013frontier} introduce void cells to generalize frontier detection in 3D, and Cieslewski et al. \cite{cieslewski2017rapid} prioritize frontiers within the sensor’s field of view while minimizing velocity discontinuities.

Most of these methods follow greedy strategies that select frontiers with the highest immediate utility. While computationally efficient, they often suffer from poor global planning, causing backtracking and suboptimal coverage. To address this, recent studies incorporate coverage path planning (CPP), which generates obstacle-aware full-coverage trajectories \cite{galceran2013survey}. Song et al. \cite{song2018surface, song2020online} propose a hybrid approach combining global surface partitioning and local viewpoint sampling for accurate 3D reconstruction, but at the cost of slower flight. Zhou et al. \cite{zhou2023racer} extend FUEL \cite{zhou2021fuel} by introducing hierarchical grids (hgrids) for multi-UAV coordination, though uniform space division may overlook obstacles and lead to inefficient paths. Zhao et al. \cite{zhao2023autonomous} refine frontier selection by considering their relation to the exploration boundary. These advancements highlight the growing focus on scalable and efficient exploration in autonomous UAV systems.

Despite the progress in both sampling-based and frontier-based exploration, many existing methods struggle to generalize across large-scale, unstructured, and elevation-rich environments. In particular, frontier-based approaches in 3D often suffer from high computational costs due to volumetric processing, as well as inefficient motion caused by the presence of blind voxels and uneven terrain. While recent efforts have introduced global planning strategies to improve coverage, they typically rely on full 3D voxel maps and require predefining the vertical extent of the environment, which limits adaptability in outdoor or unknown-scale scenarios.

\section{Map Framework}
\label{Sec: map_framework}
Compared to 3D voxel-based methods, 2D maps offer faster updates, lower memory use, and simpler planning, making them suitable for large-scale or real-time exploration. However, they lack elevation, and fixed-height projections limit adaptability in uneven terrain. While leveraging the lightweight 2.5D mapping framework proposed in \cite{Fred2024voxel}, which combines 2D efficiency with 3D spatial awareness, we further improve it by addressing the height bootstrapping issue encountered during exploration. This helps the robot explore at the correct height. The mapping process is illustrated in the map module of Fig. \ref{fig: framework}.

\subsection{Extraction of Free Space and Height Map}
\label{Sec: extraction_free_space}
We begin by extracting key spatial information from the 3D voxel map $\mathcal{V}$, built using UFOMap \cite{duberg2020ufomap} (Fig. \ref{fig: geometry_maps}(a)), to generate a 2.5D map representation following \cite{Fred2024voxel}. As an extension of OctoMap \cite{hornung2013octomap}, UFOMap enhances traversal by storing child occupancy flags in each octree node and uses Morton codes for fast indexing and spatial queries. To improve adaptability in unknown environments, the floor is detected based on free space rather than occupied voxels.

\subsubsection{Free and Occupied Space} The 3D map $\mathcal{V}$ is projected onto a 2D grid $\mathcal{G}_{2D}$ of size $M \times N$, where each cell $\{x_m, y_n\}$ contains both free and occupied vertical segments, denoted by $\mathcal{F}$ and $\mathcal{O}$, respectively. 
A free-space set $F_{x_m,y_n} = \{f_1, f_2, \ldots, f_{k_{mn}}\}$ consists of disjoint height intervals $f_i = \{\check{f}_i, \hat{f}_i\}$, where $\check{f}_i$ and $\hat{f}_i$ represent the lower and upper bounds along the vertical axis. 
Similarly, the occupied-space set $O_{x_m,y_n} = \{o_1, o_2, \ldots, o_{l_{mn}}\}$ defines occupied height intervals $o_i = \{\check{o}_i, \hat{o}_i\}$. 
Free ranges smaller than a robot-dependent threshold $R_{\text{thresh}}$ are filtered out to ensure sufficient clearance for navigation in the projected 2D map.

\begin{figure}[t]
    \centering
    \includegraphics[width=\columnwidth]{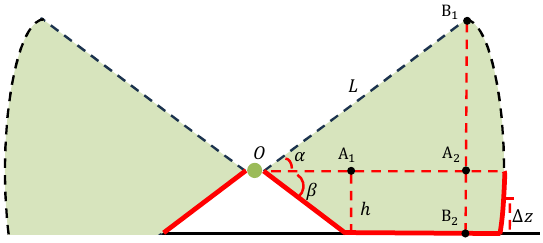}
    \caption{Confidence analysis of the floor estimation based on the LiDAR sensing model.}
    \label{fig: height_bootstrapping}
\end{figure}

\subsubsection{Height Map} From the remaining free-space intervals, a height map $\mathcal{H} = \{h_{x_m, y_n}\}_{M \times N}$ is generated, where each element $h_{x_m, y_n} = \{\check{h}_{x_m, y_n}, \hat{h}_{x_m, y_n}\}$ defines the estimated floor and ceiling heights. The floor $\check{h}_{x_m, y_n}$ corresponds to the bottom of the first free interval $f_1$, while the ceiling $\hat{h}_{x_m, y_n}$ is defined as the top of the first free interval instead of the last one. This design accounts for unstructured outdoor environments such as forests, where multiple layers of navigable space may exist. This differs from the assumption made in \cite{Fred2024voxel}, which considers only a single navigable layer. These height values reflect local terrain elevation and roughness, as shown in Fig. \ref{fig: geometry_maps}(b).

\subsubsection{Sensor Model Correction}
However, this floor estimation method is inherently greedy and may misidentify the actual ground in certain scenarios, particularly when the robot's altitude changes or near the boundary of the LiDAR’s sensing range, as illustrated in Fig. \ref{fig: height_bootstrapping}. If such cases are not properly handled, they can lead to height bootstrapping during exploration, potentially disrupting the robot's ability to explore the environment effectively. Therefore, we refine the construction of the height map by incorporating the LiDAR sensor model, enabling more robust ground estimation under these conditions, following
\begin{equation}
\begin{aligned}
    O\!A_1 &= \frac{h}{tan\beta} \\
    O\!A_2 &= Lcos\alpha \\
    A_1\!A_2 &= Lcos\alpha - \frac{h}{tan\beta}
\end{aligned}
\end{equation}
here, $\Delta z$ denotes the vertical offset between the actual ground surface and the LiDAR points near the edge of the sensor's range. During exploration, the UAV continuously moves toward the boundary between known and unknown regions, and its motion is influenced by underlying terrain changes (see Sec. \ref{Sec: frontier_detection_and_clustering}). As a result, these edge-related errors tend to accumulate throughout the exploration process. This causes a gradual increase in the UAV's altitude, leading to the height bootstrapping issue mentioned earlier. Point $A_1$ denotes the initial LiDAR return on the ground surface, while $A_2$ represents a conservatively selected point marking the end of the valid ground return interval. The segment $A_1\!A_2$ thus defines the effective ground observation range of the LiDAR. Since our system operates in an exploration setting, inaccuracies in near-range height estimation have minimal impact on distant exploration. To improve ground coverage, we extend the effective range and use the region $O\!A_2$ to update the height map. To ensure that $A_1\!A_2$ remains within the valid sensing range, we constrain the flight altitude as follows:
\begin{equation}
    B_1\!B_2 \leq L(sin\alpha+sin\beta)
\end{equation}

\subsection{Generation of 2D Occupancy Maps and Slope Map}
\label{Sec: map_generation}
\begin{figure}[t]
    \centering
    \includegraphics[width=\columnwidth]{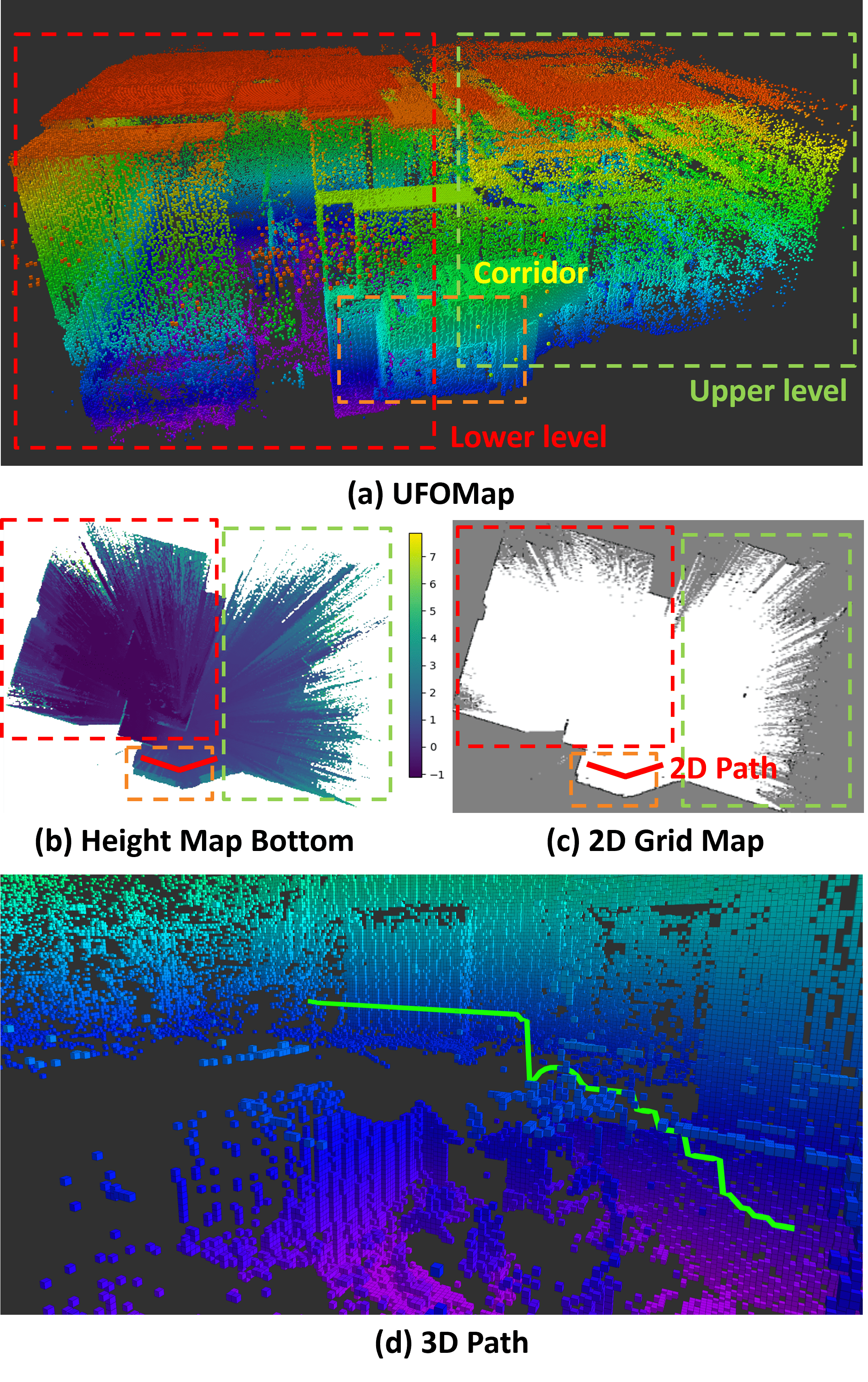}
    \caption{Illustration of our 2.5 map framework with embedded 3D spatial information and the corresponding 3D path generation process. (a) The UFOMap captures a multi-level environment with significant elevation differences, including a corridor between upper (green box) and lower (red box) levels. (b) The bottom surface of the Height Map encodes terrain elevation. (c) The 2D Grid Map supports efficient frontier detection and path planning, where the red path is initially planned in the 2D space. (d) The 2D path is lifted into a 3D path that follows elevation changes and ascends through the corridor to the upper level.}
    \label{fig: geometry_maps}
\end{figure}

Cells with non-empty free-space sets ($F_{x_m,y_n} \neq \varnothing$) are labeled as free ($m_{x_m,y_n} = 0$), while the occupancy probability of remaining cells is computed using the following formulation:
\begin{equation}
    \label{equ: generate_2d_map_1}
    m_{x_m,y_n} =
    \begin{cases} 
    O_{cc}(x_m, y_n) & \text{if } O_{cc}(x_m, y_n) \geq O_{min} \\
    -1 & \text{else}
    \end{cases}
\end{equation}
where \( O_{cc}(x_m, y_n) \) represents the occupancy value for the examined map cell \( m_{x_m,y_n} \), ranging from 0 to 1, and \( O_{min} \) denotes the minimum allowable occupancy value. The resulting 2D grid map is illustrated in Fig. \ref{fig: geometry_maps}(c).

A least-squares plane is fitted to the neighboring cell heights $\{x_i, y_j\}$ within a configurable window $S_A$ to estimate the local floor slope at each cell, following
\begin{equation}
\underbrace{
\begin{bmatrix}
x_{1} & y_{1} & 1 \\
x_{2} & y_{2} & 1 \\
\vdots & \vdots & \vdots \\
\end{bmatrix}
}_{A}
\begin{bmatrix}
a \\
b \\
c
\end{bmatrix}
=
\underbrace{
\begin{bmatrix}
\check{h}_{x_{1}, y_{1}} \\
\check{h}_{x_{2}, y_{2}} \\
\vdots \\
\end{bmatrix}
}_{B}
\end{equation}

\begin{figure}[t]
    \centering
    \includegraphics[width=\columnwidth]{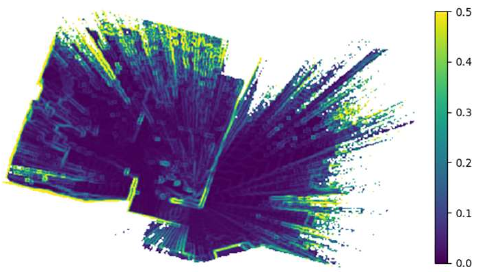}
    \caption{Visualization of the generated slope map, where each grid cell encodes the local terrain steepness. Brighter regions indicate steeper slopes, while darker regions correspond to flatter areas. This map is used to evaluate traversability and guide smoother UAV exploration planning. For visualization purposes, the slope values are capped at a maximum of 0.5 to enhance contrast.}
    \label{fig: path_maps}
\end{figure}

The slope value at each cell $\{x_m, y_n\}$ is then computed as $s_{x_m, y_n} = \sqrt{a^2 + b^2}$, where $a$ and $b$ represent the planar gradients along the $x$ and $y$ axes, respectively. The slope map is visualized in Fig. \ref{fig: path_maps}. For additional implementation details and parameter definitions, readers are referred to \cite{Fred2024voxel}.

\subsection{2D to 3D Path Conversion}
\label{Sec: 2d_to_3d_path_conversion}
Path planning is performed on a 2D grid instead of a full 3D voxel map, which can reduce computational overhead. To account for elevation, we lift the 2D path into 3D using the height map. A collision-free 2D path \( \mathcal{P} \) is first generated and then lifted using the sliding method, a greedy yet fast and effective approach. For each point \( \mathcal{P}_i \), its corresponding $Z$ coordinate is computed using:
\begin{equation}
\begin{aligned}
L_i &= \max_{j \in [i - p_f,\, i + p_f]} \check{\mathcal{H}}(\mathcal{P}_j) + R_{\text{off}} \\
U_i &= \min_{j \in [i - p_f,\, i + p_f]} \hat{\mathcal{H}}(\mathcal{P}_j) - R_{\text{clear}} \\
\mathcal{P}_{iZ} &= \mathrm{clip}(L_i, U_i)
\end{aligned}
\end{equation}
where $\check{\mathcal{H}}(\mathcal{P}_j)$ and $\hat{\mathcal{H}}(\mathcal{P}_j)$ denote the floor and ceiling values at location $\mathcal{P}_j$ from the height map $\mathcal{H}$. To ensure safe navigation, $R_{\text{off}}$ specifies the desired clearance above the ground, while $R_{\text{clear}}$ defines the minimum vertical distance to the ceiling. The parameter $p_f$ specifies the forward and backward steps used to compute the height window. Unlike \cite{Fred2024voxel}, we use the ceiling of the first free interval $f_1$, combined with a sliding-window minimum to suppress height fluctuations. This helps the path anticipate overhead occlusions and adapt to changes in complex, unstructured outdoor environments. Clearly, the thresholds $R_{\text{off}}$, $R_{\text{clear}}$, and $R_{\text{thresh}}$ are configured to satisfy the following condition:
\begin{equation}
    R_{\text{thresh}} > R_{\text{off}} + R_{\text{clear}}
\end{equation}

Our framework supports a wide range of 2D path planners; A* is adopted in this implementation. The computed 2D path (Fig. \ref{fig: geometry_maps}(c)) is subsequently extended into a 3D path using floor elevation data, as shown in Fig. \ref{fig: geometry_maps}(d).

\section{Exploration Planning}
3D voxel maps are widely used in exploration due to their ability to retain detailed environmental structures \cite{zhou2021fuel, huang2023fael}. However, they often suffer from high computational overhead in frontier detection and limited adaptability to terrain variations, particularly due to fixed vertical bounds and the presence of blind voxels. To address these limitations, we execute a 3D exploration planning based on a 2.5D map framework, as introduced in Sec. \ref{Sec: map_framework}. The detailed planning process is illustrated in the exploration planning pipeline of Fig. \ref{fig: framework}.
\subsection{Blind Voxel Problem}
We define a global voxel map $\mathcal{V}$, where $\mathcal{V}_{\text{update}}$ denotes the set of voxels updated at the current time, and $\mathcal{V}_{\text{FOV}}$ represents the region within the current sensor's field of view. However, in outdoor environments, LiDAR often fails to return measurements in regions such as the sky or distant open spaces, leaving parts of the map unobserved and therefore unupdated. We refer to these voxels as blind voxels, denoted by $\mathcal{V}_{\text{blind}}$, and define them as follows:
\begin{equation}
V_{\text{blind}} = V_{\text{FOV}} \setminus V_{\text{update}}
\end{equation}

The free-range projection approach used in the map framework mentioned in Sec. \ref{Sec: map_framework} provides a potential solution to this issue. For regions within the sensor's FOV that remain unupdated, the method infers occupancy by utilizing updated voxels within the same area and their overlapping update information with surrounding regions. Furthermore, the 3D information from the bottom-most voxels is utilized to capture local height variations, allowing the framework to retain and continuously update the 3D spatial structure of the environment.

\begin{figure}[t]
    \centering
    \includegraphics[width=\columnwidth]{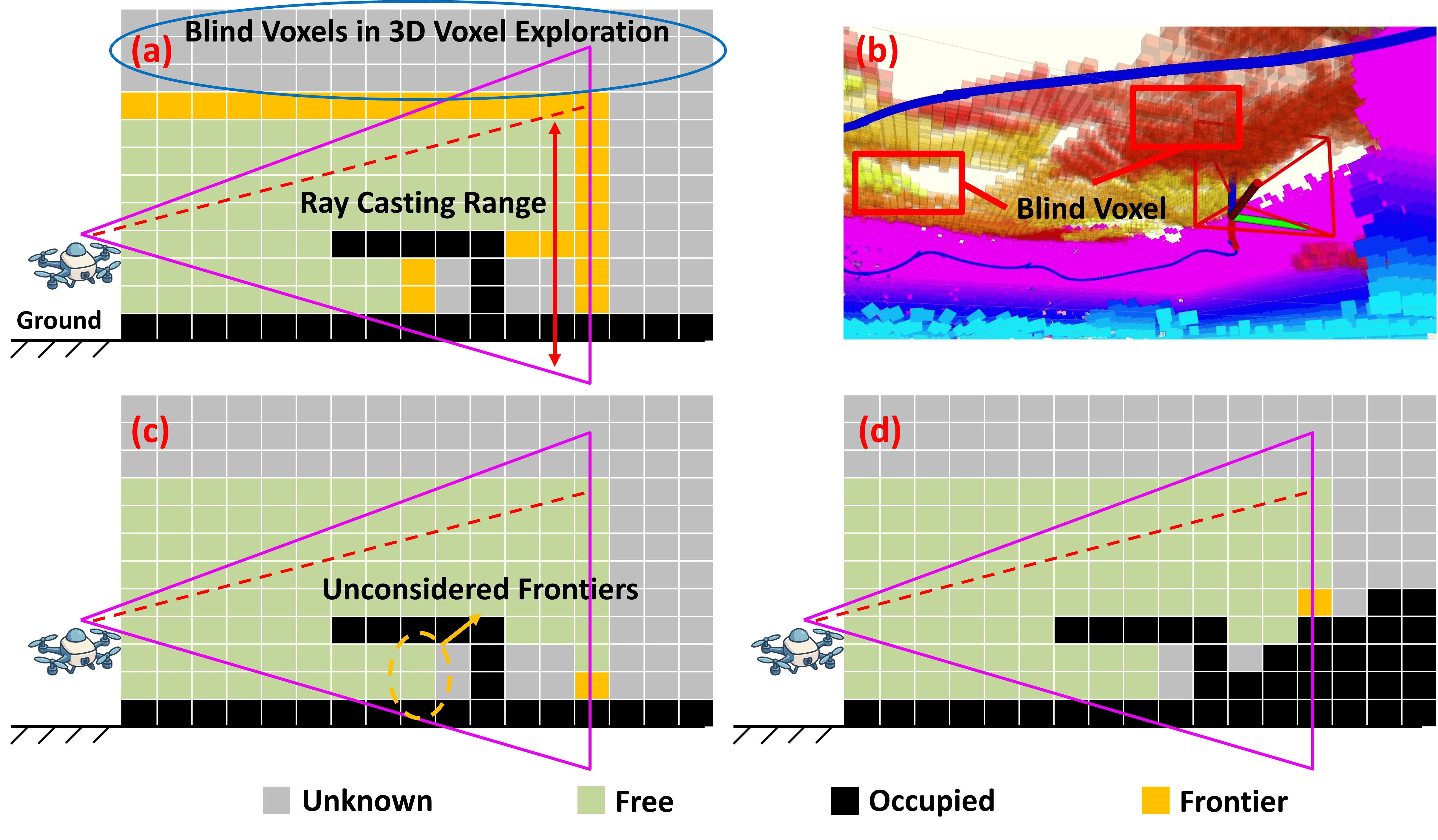}
    \caption{Comparison of conventional frontier detection limitations caused by blind voxels and the improvements achieved by our method through a 2.5D map framework. (a) Simulation of blind voxels caused by missing sensor returns in certain directions results in persistent unobserved voxels in mid-air. (b) Real-world visualization of blind voxels, which can degrade exploration efficiency and cause the UAV to loiter. (c) Detected frontiers using our method under the same scenario. Regions with a free range below the safety threshold are excluded from frontier detection. (d) Frontier height adapts to the elevation variations in rugged terrain.}
    \label{fig: blind_voxel}
\end{figure}

\subsection{Frontier Detection and Clustering}
\label{Sec: frontier_detection_and_clustering}
A key advantage of the 2D grid map in frontier detection is reduced dimensionality. Traditional 3D frontier detection requires checking 26 neighboring voxels in a $3\times3\times3$ neighborhood (excluding the center voxel) to determine adjacency to unknown space, whereas 2D detection only examines 8 neighboring cells, resulting in simpler and faster computation. With fewer elements and reduced complexity, 2D frontier detection enables faster updates, particularly in large-scale environments or when using wide field-of-view sensors.

The 2D maps are incrementally updated based on changes in UFOMap. To accelerate frontier updates, the updated axis-aligned bounding box (AABB) from UFOMap is used to localize potential frontier changes. Frontiers in these regions are then detected and clustered using a region-growing algorithm on the 2D grid map $\mathcal{M}$. If the number of frontiers exceeds a predefined threshold, they are grouped into clusters for further processing. For each frontier $\{x_m, y_n\}$, the corresponding elevation is retrieved from the bottom layer of the height map $\mathcal{H}$ as $\check{h}_{x_m, y_n}$. Oversized clusters are recursively split into balanced sub-clusters using Principal Component Analysis (PCA), with each assigned its own AABB. A detailed update is only triggered when the updated AABB intersects a cluster’s AABB. Changed clusters are discarded, and new frontiers are re-detected within the affected area. Since all operations are carried out in the 2D space, the proposed two-stage approach, which first detects frontiers in 2D and then retrieves their corresponding heights, is more efficient than directly performing frontier detection in 3D voxel space. The computational advantage can be expressed as follows:
\begin{equation}
    \frac{T_{3D}}{T_{2D}} = \frac{O(N_{xy}H)}{O(N_{xy})} = O(H)
\end{equation}
where $T_{3D}$ and $T_{2D}$ are the computation times for 3D and 2D frontier detection. $N_{xy}$ is the number of horizontal grid cells, and $H$ is the number of vertical layers. The result indicates that 3D detection is $O(H)$ times more expensive than 2D.

In 3D voxel-based maps, memory is typically pre-allocated to cover a fixed vertical (z-axis) extent, which limits the system’s ability to adapt to varying terrain elevations during exploration. An overly large vertical range increases memory usage and creates many blind voxels, while a small range risks omitting key areas, leading to incomplete mapping. To address this, we use a height map that dynamically adapts to terrain changes without relying on predefined height constraints. As shown in Fig. \ref{fig: blind_voxel}(c) and (d), frontiers are extracted at varying elevations according to terrain shape. The system also filters out points below a safety threshold, preventing the UAV from navigating into areas with insufficient clearance (Fig. \ref{fig: blind_voxel}(c)).

Outdoor environments often contain blind voxels caused by occlusions, limited sensor returns, or open terrain with sparse vertical structures \cite{wang2024whole}. In such cases, LiDAR points concentrate near the ground, leaving many unknown voxels overhead that cannot be observed from any angle. These blind voxels, illustrated in Fig. \ref{fig: blind_voxel}(a) and (b), artificially inflate exploration indicators in already-visited areas, prompting redundant UAV behavior. As seen in Fig. \ref{fig: blind_voxel}(b), the UAV may hover repeatedly in an attempt to clear unreachable mid-air regions. Our method avoids this inefficiency by excluding frontiers aimed at blind voxels. As shown in the comparison between Fig. \ref{fig: blind_voxel}(a) and (c), we focus only on ground-level frontiers, significantly improving exploration efficiency in outdoor terrain.

\subsection{Viewpoints Generation and Cost Update}   
\label{Sec: viewpoints_generation}
Due to the vertical mounting of the LiDAR, we omit viewpoint sampling and instead use the average position of each cluster directly as the target viewpoint. This approach removes the need for volumetric gain computation, which significantly reduces computational load \cite{schmid2020efficient}. To ensure safe flight, the UAV must maintain a proper clearance from the ground. Therefore, the height of each viewpoint in the set $\text{VP} = \{\mathbf{x}_1, \mathbf{x}_2, ..., \mathbf{x}_n\}$ is determined as follows:
\begin{equation}
    x_{i}^{z} = x_{i}^{z} + R_{safe}
\end{equation}
where $R_{safe}$ is a user-defined safety distance.

One of the key advantages of our approach is that it enables viewpoint-to-viewpoint cost computation using 2D path planning mentioned in Sec. \ref{Sec: 2d_to_3d_path_conversion}, rather than relying on full 3D voxel-based. This improves planning efficiency in large-scale 3D environments. We compute the time lower bound $t_{lb}(\mathbf{x}_i, \mathbf{x}_j)$ when moving between two viewpoints by:
\begin{align}
t_{\mathrm{lb}}(\mathbf{x}_i, \mathbf{x}_j)
= &
    \frac{\operatorname{length}\left( P(\mathbf{p}_i, \mathbf{p}_j) \right)}{v_{\max}}
\end{align}
where $P(\mathbf{p}_{i}, \mathbf{p}_{j})$ denote a collision-free 3D path between $\mathbf{p}_{i}$ and $\mathbf{p}_{j}$ found by a path searching algorithm based on both 2D grid map and height map, $v_{\text{max}}$ is the limits of velocity. The resulting time lower bound reflects the geometry-based cost.

\begin{figure}
    \centering
    \includegraphics[width=\columnwidth]{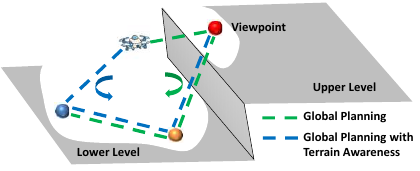}
    \caption{Illustration of terrain-aware global planning. The green dashed line shows a globally planned path that does not consider terrain variation, resulting in two steep climbs between levels. In contrast, the blue dashed line incorporates terrain information and follows a smoother route with only one elevation transition.}
    \label{fig: terrain_awareness}
\end{figure}

Another advantage of our method is its ability to reduce inefficiencies caused by UAVs oscillating between viewpoints located on uneven terrain. After computing the geometry-based cost using the 2D grid map and height map, we incorporate slope map information to evaluate the smoothness of terrain transitions between viewpoints. As illustrated in Fig. \ref{fig: terrain_awareness}, relying solely on geometric cost can lead the UAV to select viewpoints that appear optimal but involve steep elevation changes, which may result in unexpected altitude shifts during exploration. By adding terrain-aware cost, the planning encourages smoother transitions and helps maintain spatial consistency in the overall exploration trajectory.

Since the 2D grid map, height map, and slope map are all aligned within the same coordinate frame, we can directly retrieve the slope $s$ for each waypoint along $P(\mathbf{p}_i, \mathbf{p}_j)$. These slope values are computed by fitting a local plane within a predefined area $S_A$, which ensures that traversability evaluation reflects both point-wise terrain features and broader terrain trends along the path corridor. In our framework, a 2D path is considered feasible if it lies entirely within a connected traversable region that also exceeds a minimum spatial threshold. Once the slope set $\{s_1, s_2, ..., s_n\}$ is obtained for the entire path, we evaluate terrain traversability by computing a local traversability score $\tau_i$ for each waypoint $P_i$ as follows:
\begin{equation}
\tau_i =
\begin{cases}
\left(\dfrac{s_i}{s_{\text{crit}}}\right)^2, & s_i < s_{\text{crit}} \\
C, & s_i \geq s_{\text{crit}}
\end{cases}
\end{equation}
where $s_{crit}$ is the predefined slope threshold, and $C$ is a high penalty value used to discourage paths involving abrupt elevation changes beyond this threshold. The overall traversability of a path is computed as the sum of the local traversability at each point along the path:
\begin{equation}
\mathcal{T}_{P}(\mathbf{x}_i, \mathbf{x}_j) = \sum_{i=1}^{N} \tau_i
\end{equation}
where $\mathcal{T}_{P}(\mathbf{x}_i, \mathbf{x}_j)$ denotes the terrain-aware cost between two viewpoints.

\subsection{Terrain Awareness Global Tour Planning}
To efficiently plan a global tour covering all clusters, similar to \cite{zhou2021fuel}, we formulate the problem as the Asymmetric Traveling Salesman Problem (ATSP), starting from the current viewpoint. We construct a cost matrix $\mathbf{M}_{tsp} \in \mathbb{R}^{(|V\!P|+1) \times (|V\!P|+1)}$, where the first row/column corresponds to the UAV’s position, and the rest to the clusters. The inter-cluster costs are precomputed using a lower-bound time estimate and a terrain-aware cost between viewpoint pairs:
\begin{equation}
\begin{aligned}
    \mathbf{M}&_{tsp}(i, j) = \mathbf{M}_{tsp}(j, i) \\ &= t_{lb}(\mathbf{x}_i, \mathbf{x}_j) + w_t \cdot \mathcal{T}_{P}(\mathbf{x}_i, \mathbf{x}_j)
\end{aligned}
\end{equation}
where $w_t$ denotes the weight for the terrain-aware cost. From current position $\mathbf{x}_0$, the cost to the k-th cluster is evaluated by:
\begin{equation}
    \mathbf{M}_{tsp}(0, k) = t_{lb}(\mathbf{x}_0, \mathbf{x}_{k}) + w_t \cdot \mathcal{T}_{P}(\mathbf{x}_0, \mathbf{x}_k) + w_c \cdot c_c(\mathbf{x}_{k}) 
\end{equation}
where $w_c$ denotes the weight for the motion consistency cost. The cost is designed to discourage frequent direction reversals in successive planning steps as follows:
\begin{equation}
    c_c(\mathbf{x}_{k}) = \cos^{-1} \left( \frac{(\mathbf{p}_{k} - \mathbf{p}_0) \cdot \mathbf{v}_0}{\lVert \mathbf{p}_{k} - \mathbf{p}_0 \rVert \lVert \mathbf{v}_0 \rVert} \right)
\end{equation}
where $\mathbf{v}_0$ is the current velocity. To convert the open-loop TSP to a standard ATSP solvable by existing solvers, we assign zero cost from each cluster back to the current position:
\begin{equation}
    \mathbf{M}_{tsp}(k, 0) = 0, \quad \forall k
\end{equation}

This ensures that the optimal closed-loop solution contains an equivalent open-loop path without additional cost.

\begin{table}[t]
    \centering
    \caption{Parameter Settings}
    \label{tab: parameters}
    \resizebox{0.8\columnwidth}{!}{
    \begin{tabular}{llll}
    \hline
    \textbf{Parameter} & \textbf{Value} & \textbf{Paramater} & \textbf{Value} \\ \hline
    $R_{thresh}$ & 2.0 & $O_{\text{min}}$ & 0.5 \\
    $R_{of\!f}$ & 1.0 & $R_{clear}$ & 0.5 \\
    $R_{\text{safe}}$ & 1.5 & $S_A$ & 4 \\
    $p_f$ & 15 & $s_{crit}$ & 0.6 \\
    $C$ & 4 & $w_t$ & 0.8 \\
    $w_c$ & 1.3 & Resolution & 0.1  \\ \hline
    \end{tabular}
    }
\end{table}

\section{Experiments}
\subsection{Simulation Benchmark Results And Analysis}
\label{Sec: simulation_benchmark}
In this section, we conducted a comprehensive comparative analysis between the proposed method and five state-of-the-art exploration algorithms, including FUEL \cite{zhou2021fuel}, FAEP \cite{zhao2023autonomous}, NBVP \cite{bircher2016receding}, RACER \cite{zhou2023racer}, and frontier-based exploration using our 2.5D map (FBE) \cite{yamauchi1997frontier}. Although RACER was originally designed for multi-robot exploration, we adapted it for a single-robot configuration to ensure a fair and consistent benchmark across all methods. To ensure a fair comparison and statistical robustness, all algorithms were tested in the same simulation environment, with four independent runs per scenario under identical initial conditions. The parameters of the proposed framework are summarized in Table \ref{tab: parameters}. Since the vertical extent of the real-world environment is usually unknown, we provided this information only in simulation to all baseline methods to eliminate any advantage from map representation differences. The sensing configuration was standardized using a simulated MID360 LiDAR with a maximum range of 20 m. All UAVs were constrained to a maximum flight speed of 2 m/s, and each run was limited to a maximum exploration duration of 600 seconds (10 minutes).  All comparative experiments were conducted on an Intel NUC with an Intel Core i7-1260P CPU running Ubuntu 20.04. The evaluation was conducted in environments with significant elevation changes, which were reconstructed from real-world LiDAR scans and tested using the point-realistic simulation platform, MARSIM \cite{kong2023marsim}. These environments were designed to reflect the natural variability and terrain complexity commonly encountered in real-world field deployments:

\begin{figure}[t]
    \centering
    \includegraphics[width=\columnwidth]{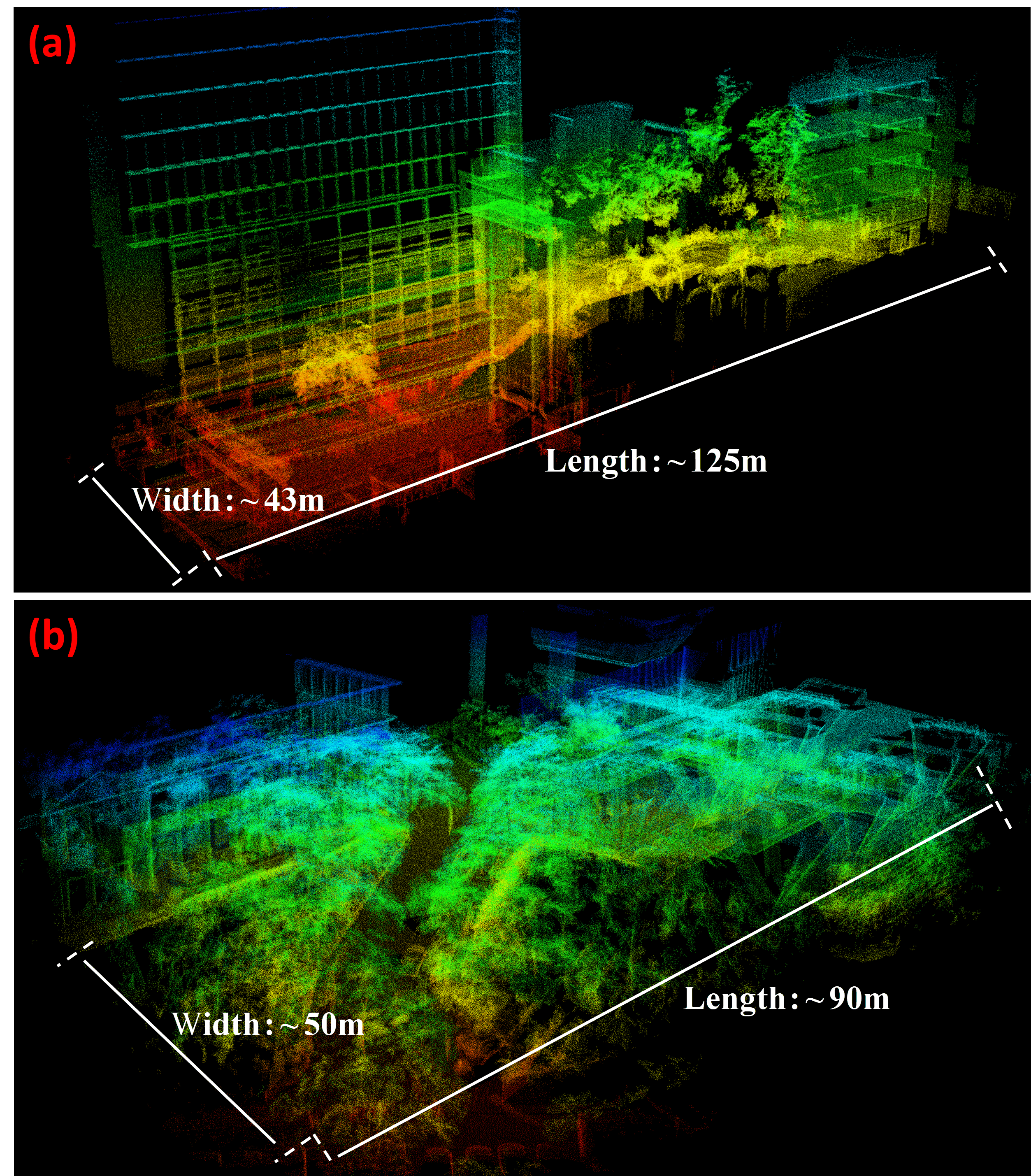}
    \caption{The visualization of the two simulation scenarios. (a) Long Corridor-Like Environment. (b) Mixed Forest-Urban Environment.}
    \label{fig: simulation_1}
\end{figure}

\begin{table*}[ht]
\centering
\caption{Exploration Performance Statistics across Different Methods and Scenarios}
\label{tab: exploration_statistics}
\resizebox{\textwidth}{!}{
\begin{tabular}{ll|cccc|cccc|cccc}
\toprule
\textbf{Scene} & \textbf{Method} 
& \multicolumn{4}{c|}{\textbf{Exploration Time (s)}} 
& \multicolumn{4}{c|}{\textbf{Flight Distance (m)}} 
& \multicolumn{4}{c}{\textbf{Avg Velocity (m/s)}} \\
& & Avg & Std & Max & Min 
& Avg & Std & Max & Min 
& Avg & Std & Max & Min \\
\midrule
\multirow{5}{*}{\makecell{Scene 1}} 
& NBVP \cite{bircher2016receding}   & 600.0 & 0.00 & 600.0 & 600.0 & 319.4 & 3.25 & 322.5 & 313.5 & 0.53 & 0.01 & 0.54 & 0.53 \\
& FUEL \cite{zhou2021fuel}     & 600.0 & 0.00 & 600.0 & 600.0 & 383.3 & 40.92 & 432.2 & 334.8 & 0.87 & 0.12 & 1.00 & 0.73 \\
& RACER \cite{zhou2023racer}   & 600.0 & 0.00 & 600.0 & 600.0 & 667.9 & 128.11 & 836.3 & 508.2 & 1.10 & 0.21 & 1.39 & 0.84 \\
& FAEP \cite{zhao2023autonomous}   & 571.3 & 34.90 & 600.0 & 529.2 & 467.9 & 76.50 & 549.2 & 364.8 & 0.84 & 0.06 & 0.92 & 0.79 \\
&FBE \cite{yamauchi1997frontier} & 286.3 & 20.21 & 302.7 & 263.7 & 396.9 & 23.16 & 419.2 & 373.0 & 1.39 & 0.06 & 1.43 & 1.31 \\
& \textbf{Ours}    & 194.6 & 19.82 & 219.6 & 172.8 & 327.4 & 37.63 & 378.8 & 287.1 & 1.68 & 0.02 & 1.71 & 1.66 \\
\midrule
\multirow{5}{*}{\makecell{Scene 2}} 
& NBVP \cite{bircher2016receding}   & 600.0 & 0.00 & 600.0 & 600.0 & 308.1 & 7.35 & 314.8 & 300.3 & 0.52 & 0.01 & 0.53 & 0.51 \\
& FUEL \cite{zhou2021fuel}     & 586.6 & 17.85 & 600.0 & 562.3 & 534.9 & 58.06 & 599.0 & 465.3 & 0.92 & 0.12 & 1.03 & 0.77 \\
& RACER \cite{zhou2023racer}   & 600.0 & 0.00 & 600.0 & 600.0 & 664.7 & 30.43 & 707.3 & 624.4 & 1.10 & 0.05 & 1.17 & 1.03 \\
& FAEP \cite{zhao2023autonomous}   & 564.7 & 50.35 & 600.0 & 493.2 & 570.6 & 24.71 & 602.6 & 547.7 & 1.01 & 0.09 & 1.12 & 0.92 \\
&FBE \cite{yamauchi1997frontier} & 198.6 & 11.38 & 210.8 & 188.2 & 307.4 & 19.23 & 329.1 & 292.4 & 1.55 & 0.03 & 1.57 & 1.52 \\
& \textbf{Ours}    & 157.5 & 14.42 & 172.3 & 137.3 & 258.4 & 21.28 & 287.2 & 227.6 & 1.65 & 0.05 & 1.70 & 1.56 \\
\bottomrule
\end{tabular}
}
\end{table*}

\begin{figure*}
    \centering
    \includegraphics[width=\textwidth]{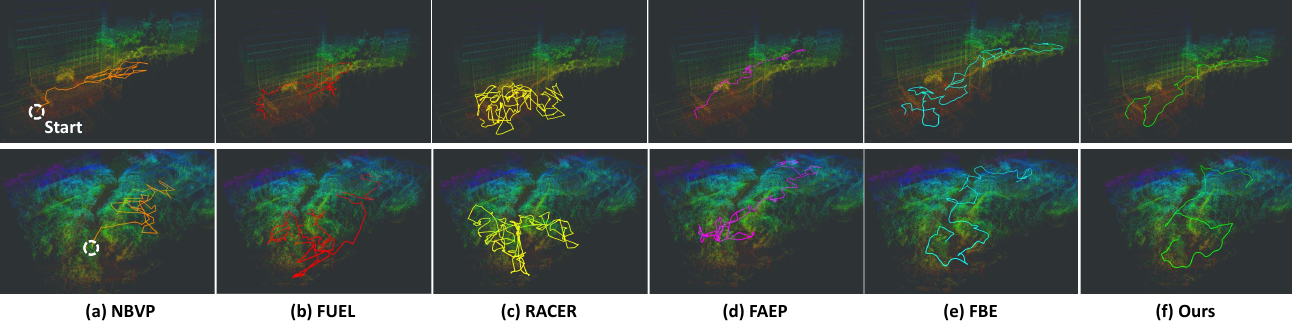}
    \caption{Comparison of exploration trajectories generated by different methods in two representative outdoor environments with significant elevation changes.}
    \label{fig: comparison_trajectory}
\end{figure*}

\begin{figure}
    \centering
    \includegraphics[width=\columnwidth]{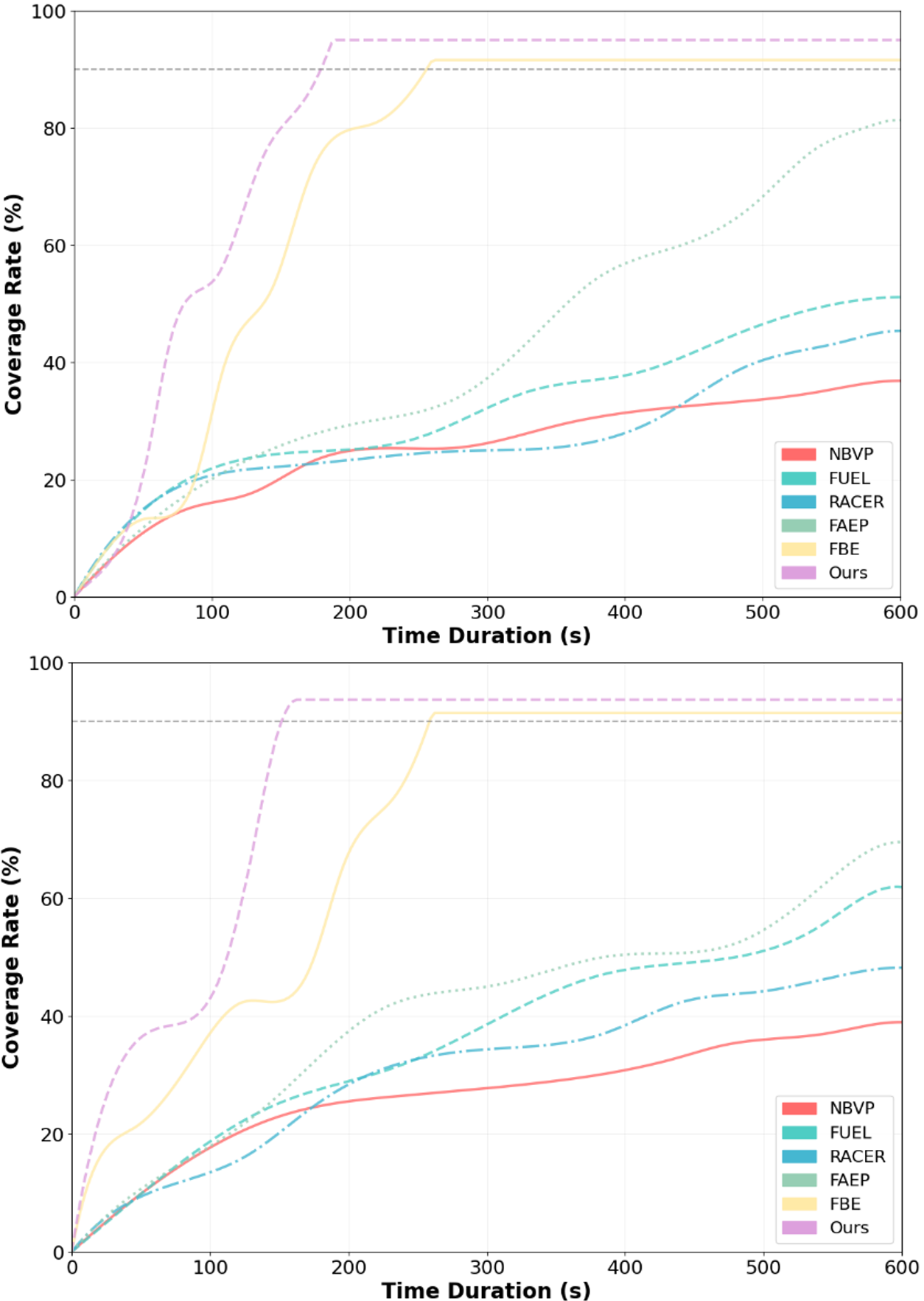}
    \caption{The coverage rate over time in Scene 1 (top) and Scene 2 (bottom).}
    \label{fig: coverage_rate}
\end{figure}

\subsubsection{Scene 1 (Long Corridor-Like Environment)}
The first scenario features a structured corridor-like environment composed of two connected square plazas at different heights, linked by a wide staircase with significant vertical variation (Fig. \ref{fig: simulation_1}(a)). This setup simulates real-world sites like campus plazas or multilevel pedestrian corridors.

\subsubsection{Scene 2 (Mixed Forest-Urban Environment)}
The second scenario presents a more unstructured environment that combines sparse forested areas with urban structures on a sloped hill (Fig. \ref{fig: simulation_1}(b)). This scene tests the ability of each algorithm to handle mixed terrain types and complex elevation profiles with limited line-of-sight.

We evaluate the exploration performance using the following three quantitative metrics:
\begin{itemize}
    \item \textbf{Exploration Time (s):} The duration required to achieve a 90\% environment coverage rate, or the maximum time limit is reached.
    \item \textbf{Flight Distance (m):} The total path length traveled by the UAV during the exploration.
    \item \textbf{Avg Velocity (m/s):} The mean speed of the UAV throughout the exploration process.
    \item \textbf{Coverage Rate (\%):} The percentage of the traversable area, obtained from our map conversion, that has been successfully covered relative to the entire environment.
\end{itemize}

\begin{table}[t]
\centering
\caption{Run Time Comparison}
\label{tab: runtime_comparison}
\resizebox{\columnwidth}{!}{%
\begin{tabular}{lcccccc}
\Xhline{1pt}
\multirow{2}{*}{Scene} & \multicolumn{6}{c}{\makecell{Methods Average Run Time (s)}} \\
\cmidrule{2-7}
& NBVP & FUEL & RACER & FAEP & FBE & Ours \\
\Xhline{0.8pt}
Scene 1 & 8.290 & 0.407  & 0.312 & 0.381 & 0.087 & 0.124 \\
Scene 2 & 9.272 & 0.722  & 0.513 & 0.492 & 0.091 & 0.147 \\
\Xhline{1pt}
\end{tabular}
}
\end{table}

The top row of Fig. \ref{fig: comparison_trajectory} illustrates the exploration trajectories generated by each algorithm in Scene 1, highlighting significant differences in trajectory structure and area coverage behavior. Our method demonstrates a clear advantage in structured environments with unknown elevation. By leveraging the 2D Height Map, which continuously provides ground elevation information, our method accurately extracts terrain-level frontiers. This allows the UAV to focus its exploration efforts on accessible surfaces while ignoring blind voxels suspended in mid-air. As a result, the UAV maintains a consistent trajectory on the current terrain surface, leading to smoother motion and improved exploration efficiency. This behavior is visually confirmed in Fig. \ref{fig: comparison_trajectory}(f), where the UAV progresses through multi-level areas without redundant hovering or detouring. 

Quantitative results in Table \ref{tab: exploration_statistics} further validate this advantage. Our method achieved full coverage in only 194.6 seconds, maintaining a high average speed of 1.68 m/s. In contrast, NBVP, FUEL, and RACER all failed to complete coverage within the time limit. Although FAEP eventually finished the task, it required nearly three times as long as our method. The poor performance of NBVP can be attributed to its long decision-making latency and lack of global trajectory planning. These limitations lead to inefficient local exploration and frequent redundant movements. Consequently, NBVP recorded the shortest flight distance (319.4 m) and the lowest average speed (0.53 m/s) among all compared methods. While FUEL, RACER, and FAEP incorporate global planning strategies, their trajectories still show noticeable local detours and inefficiencies. Among them, RACER traveled the longest distance (667.9 m) and maintained a relatively high average speed of 1.10 m/s, yet its exploration path appeared overly convoluted and spatially constrained. This is likely due to its design, which emphasizes detailed local area clearing, often at the expense of global efficiency. Additionally, in unstructured outdoor environments, the presence of blind voxels severely impacts RACER's performance, resulting in increased hesitation and redundant motion. In contrast to these methods, FBE adopts a greedy exploration strategy but, when combined with our 2.5D mapping framework, is able to efficiently navigate unstructured outdoor 3D environments. Compared to FAEP, FBE reduced the total exploration time by 50\% and the traveled distance by 15\%, while maintaining a consistently higher exploration speed. These results highlight the effectiveness of our 2.5D map representation in guiding fast and efficient 3D exploration. Fig. \ref{fig: coverage_rate} illustrates the coverage progression over time for each method. With the aid of our mapping framework, the UAV is able to rapidly cover unknown environments. Compared to FBE, our approach achieves faster coverage by incorporating global planning, resulting in a 32\% reduction in exploration time and an 18\% decrease in flight distance.

Finally, Table \ref{tab: runtime_comparison} shows that our method also offers a significant computational advantage. In terms of average runtime, it is approximately 66+ times faster than NBVP, over 3+ times faster than FUEL and FAEP, and 2.5+ times faster than RACER, highlighting the efficiency of our framework. FBE exhibits a shorter runtime than our method, as it forgoes global planning and instead greedily selects the nearest cluster.

The bottom row of Fig. \ref{fig: comparison_trajectory} shows the exploration trajectories of each algorithm in Scene 2. Unlike the relatively open and structured layout of Scene 1, this scenario introduces a more demanding environment characterized by dense, unstructured obstacles, including a mix of vegetation and built structures. Such complexity presents greater challenges for both exploration coverage and motion planning stability. Despite these challenges, our method maintains robust performance. 

As reported in Table \ref{tab: exploration_statistics}, our system completed the coverage task with a total trajectory length of 258.4 m, maintaining a high average speed of 1.65 m/s and finishing in just 157.5 seconds. Compared to FUEL and FAEP, our approach reduced coverage time by 73.2\% and 72.1\%, and trajectory length by 51.7\% and 58.3\%, respectively. It also outperformed FBE, reducing exploration time and flight distance by 21\% and 16\%, respectively. These improvements are attributed to our system’s ability to generate more direct and efficient paths while minimizing redundant movements. In contrast, NBVP and RACER again failed to complete coverage within the time limit. Although RACER maintained a relatively high average speed of 1.10 m/s and covered the longest distance of 664.7 m, its trajectory was mostly constrained to a limited subset of the environment. The method struggled with fine-grained occlusions and frequent large-volume voxel updates, both of which are particularly problematic in voxel-based frameworks under cluttered conditions. These issues lead to significant planning delays and exploration inefficiencies. While voxel maps offer rich environmental detail, they incur substantial computational overhead in dense or irregular environments. The presence of irregular obstacles causes uneven frontier distributions, which in turn disrupt the planning process and make global trajectory generation less effective. Fig. \ref{fig: coverage_rate} further illustrates the coverage rate over time for each method in this scenario, highlighting our system’s ability to achieve faster and more efficient coverage in complex environments.

Table \ref{tab: runtime_comparison} shows that all methods experienced an increase in average computation time in this more complex scene. The root cause lies in persistent occlusions and irregular frontier patterns, which tend to accumulate unresolved frontier clusters and increase global planning costs over time. Our method, however, exhibited only a modest 18.5\% increase in computation time compared to its Scene 1 result. In contrast, NBVP, FUEL, RACER, and FAEP showed increases of 11.8\%, 77.4\%, 64.4\%, and 29.1\%, respectively. FBE exhibited a 10\% increase in computation time, which is likely due to the increased mapping burden in denser and more occluded environments, resulting in longer decision-making latency. Overall, our system remained significantly more efficient across all metrics. In terms of runtime, it was approximately 63+ times faster than NBVP, over 5+ times faster than FUEL, 3.5+ times faster than RACER, and more than 3+ times faster than FAEP in Scene 2. These results confirm the scalability and adaptability of our approach in handling exploration tasks in complex, real-world-like outdoor environments.

\begin{figure}[t]
    \centering
    \includegraphics[width=\columnwidth]{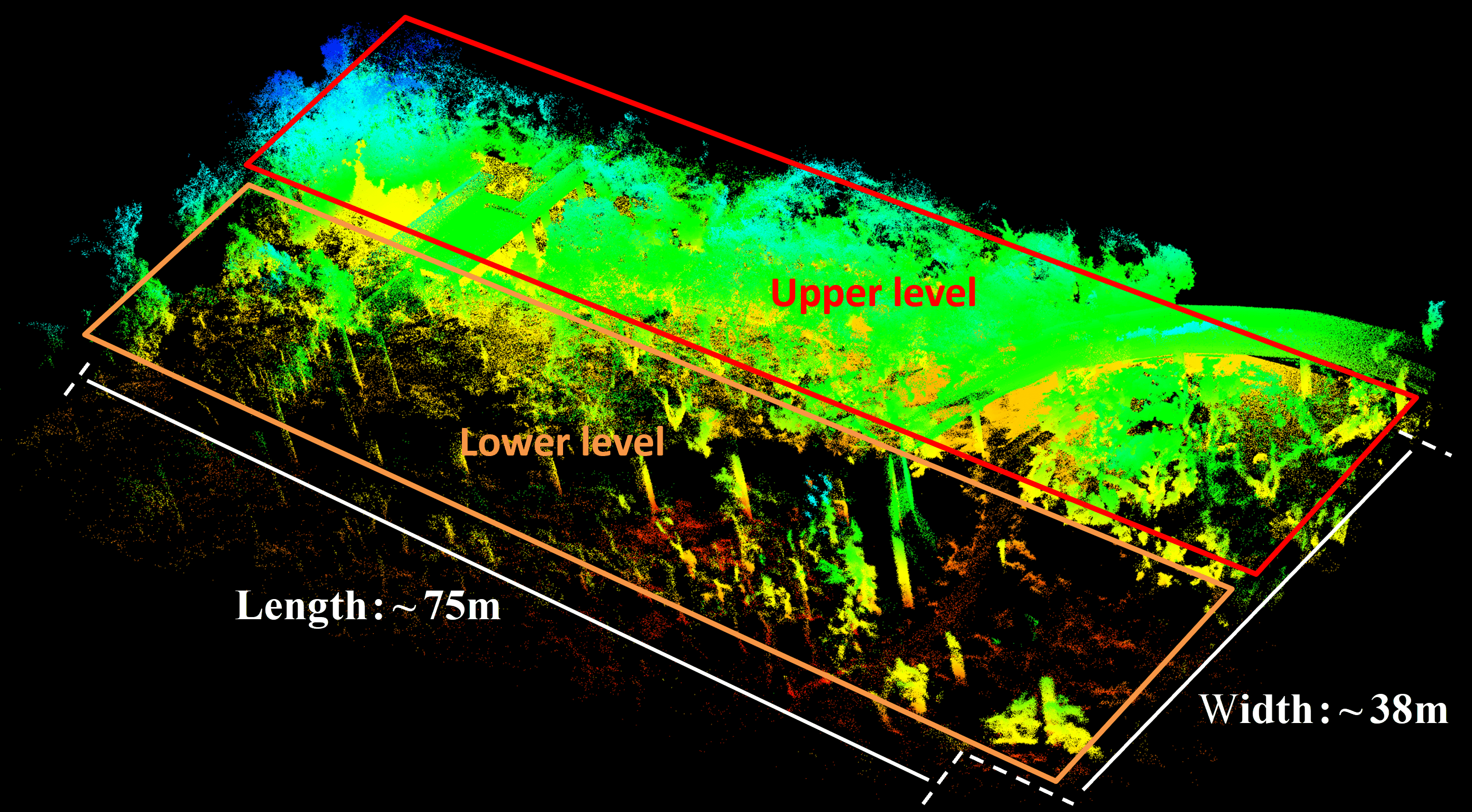}
    \caption{Overview of the complex forest environment used for evaluation. The scene features a two-level terrain structure, including an upper level and a lower level connected by forested slopes.}
    \label{fig: simulation_2}
\end{figure}

\subsection{Ablation Study}
To further validate the effectiveness of the proposed approach, we conducted a set of ablation studies using simulation experiments. Specifically, we evaluated the individual contributions of two key components in our system: the 2.5D map framework and the terrain-aware cost. All ablation tests were performed in a complex forest environment featuring a two-level terrain structure, as illustrated in Fig. \ref{fig: simulation_2}.

\subsubsection{Ablation Study of Map Framework}
\begin{figure}[t]
    \centering
    \includegraphics[width=\columnwidth]{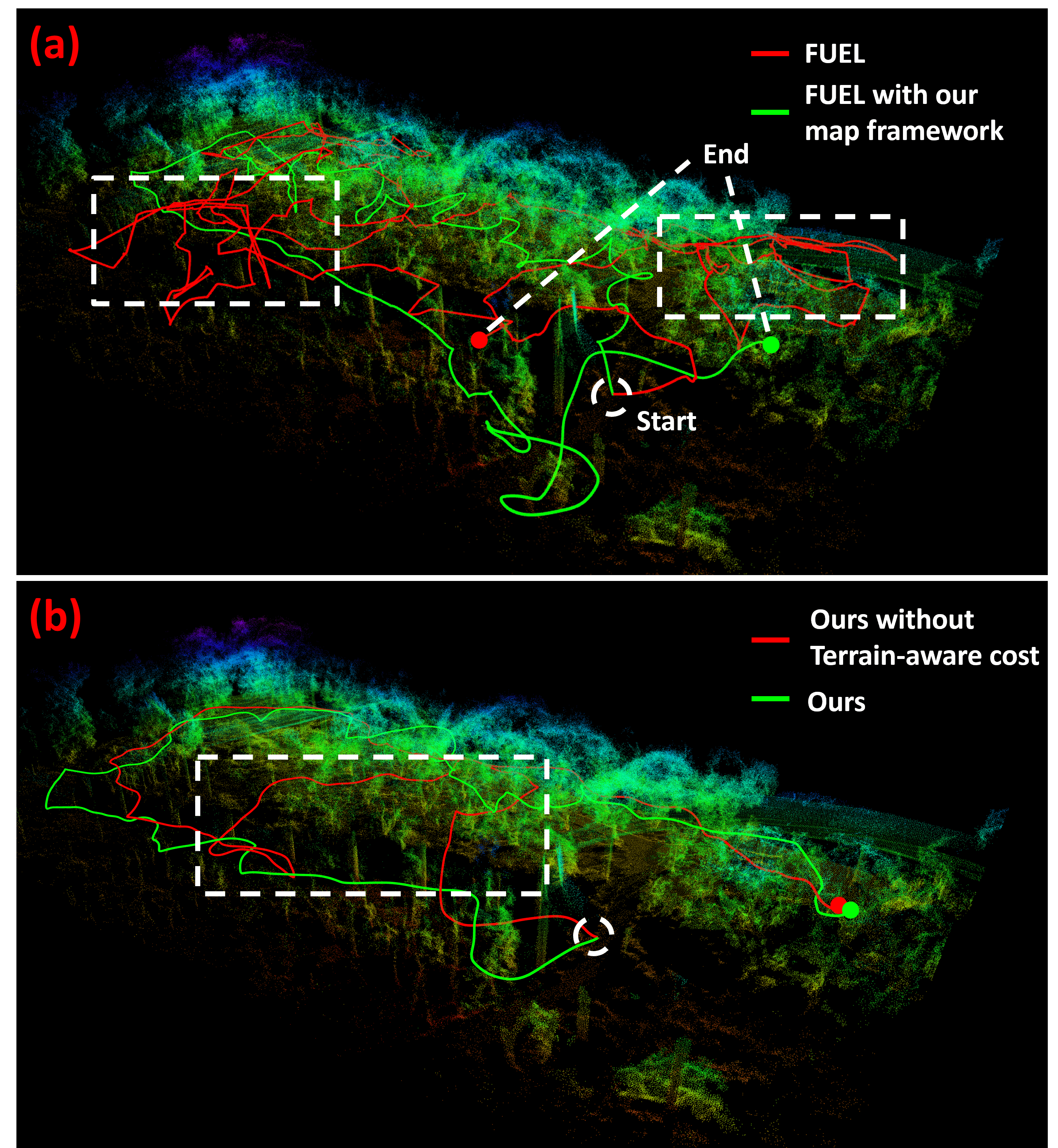}
    \caption{(a) Trajectory comparison showing the effect of replacing FUEL’s map representation with our proposed mapping framework. (b) Exploration trajectories generated by our method with and without the use of terrain traversability cost.}
    \label{fig: ablation_study}
\end{figure}

To evaluate the effectiveness of the 2.5D map framework, we replaced FUEL's original voxel-based mapping module with our 2.5D map framework while keeping all other components unchanged. The resulting performance was illustrated in Fig. \ref{fig: ablation_study}(a). This comparison highlighted the impact of blind voxel removal on exploration efficiency. After integrating our mapping framework, the exploration trajectories became significantly smoother. In contrast, the original FUEL system exhibited more convoluted trajectories due to its repeated attempts to update large volumes of blind voxels during exploration. These voxels, especially prevalent in occluded or cluttered environments, often lacked valuable information but still triggered the planner to execute inefficient back-and-forth motions in an attempt to clear them. Their non-uniform distribution further reduced planning effectiveness. By focusing solely on navigable terrain and naturally avoiding blind voxels, our map representation promoted more direct, terrain-following trajectories. As a result, FUEL's exploration time was reduced from 620 seconds to 191 seconds, and the flight distance dropped from 467 m to 284 m, corresponding to a 69\% reduction in time and a 39\% reduction in distance.

\subsubsection{Ablation Study of Terrain-aware Cost}
To assess the impact of terrain-aware cost in exploration planning, we disabled the terrain-aware cost component and compared the resulting performance, as shown in Fig. \ref{fig: ablation_study}(b). The comparison revealed that, without terrain-aware cost, the UAV tended to perform repeated altitude transitions between the two levels of the environment. These vertical shifts not only increased energy consumption but also led to frequent backtracking behavior, especially when returning to previously visited levels. This pattern highlighted the lack of vertical consistency in the executed trajectories when terrain information was ignored. With terrain awareness enabled, the planner favored smoother elevation transitions and avoided unnecessary vertical movement. This led to improved motion efficiency, as evidenced by the quantitative results. Specifically, with terrain-aware cost, the UAV completed exploration in 112.7 seconds with a total flight distance of 181.2 m. In contrast, disabling the cost resulted in a longer trajectory of 195.2 m and a slightly increased duration of 115.3 seconds. Although the time increase was modest, the additional distance indicated inefficiency in exploration due to poor terrain adaptation.

\subsection{Real-World Experiments}
\begin{figure}[t]
    \centering
    \includegraphics[width=\columnwidth]{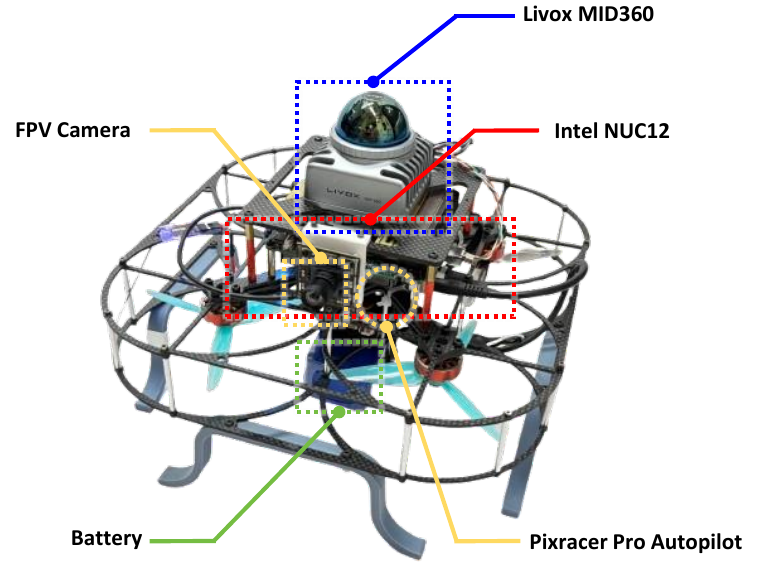}
    \caption{The quadrotor platform used in the outdoor exploration.}
    \label{fig: UAV}
\end{figure}

In the real-world experiment, the UAV platform is equipped with a Livox MID360 LiDAR for onboard perception, offering a $360^\circ$ horizontal and $59^\circ$ vertical field of view with a maximum detection range of 40 m at 10\% reflectivity. The LiDAR publishes point clouds at 10 Hz. UAV state estimation is provided by Fast-LIO2 \cite{xu2022fast}, which tightly integrates IMU prediction to achieve a 200 Hz pose update rate, enabling fast and reliable localization in challenging settings. The onboard computer is an Intel NUC with an Intel Core i7-1260P CPU running Ubuntu 20.04. All modules, including planning and localization, run fully onboard without the aid of any external infrastructure. For motion planning, we employ the SUPER \cite{ren2025safety}. The UAV dynamics are limited to a maximum velocity of $v_m =$ 5.0 $\text{m/s}$ and acceleration of $a_m = $ 5.0 $\text{m/s}^2$. An overview of the hardware setup is shown in Fig. \ref{fig: UAV}.

\begin{figure*}[t]
    \centering
    \includegraphics[width=\textwidth]{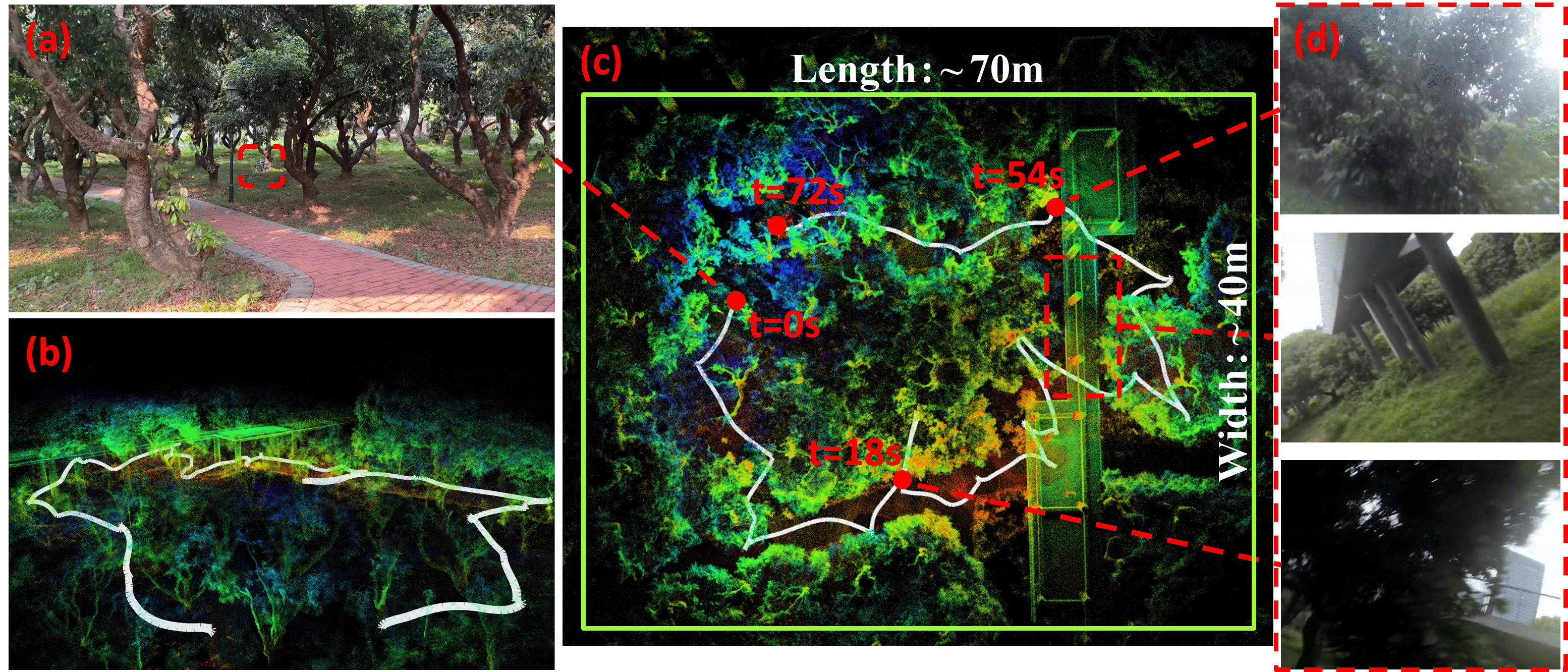}
    \caption{(a) The UAV taking off in the forest environment. (b) and (c) Two different views of the globally built online point cloud map with the executed UAV trajectory. Environment dimensions and exploration time are annotated. (d) Examples of complex obstacles encountered during exploration such as foliage and poles, along with noticeable terrain elevation changes.}
    \label{fig: exploration_result_0}
\end{figure*}

To validate the effectiveness and adaptability of our approach, we conducted an exploration mission in a real forest environment featuring a variety of natural obstacles, including grass, bushes, poles, and uneven terrain. The area spanned $70 \times 40 \times 6\ \text{m}^3$. The UAV took off from the initial position shown in Fig. \ref{fig: exploration_result_0}(a), and the online-generated map and executed trajectory were visualized from different perspectives in Fig. \ref{fig: exploration_result_0}(b) and (c). Throughout the mission, the UAV navigated through small and irregular obstacles such as low vegetation and uneven ground, which created additional challenges for fast and smooth motion, as illustrated in Fig. \ref{fig: exploration_result_0}(d). Despite the complexity of the environment, the UAV successfully completed the exploration in just 72 seconds, covering a total flight distance of 206.48 m. The executed trajectory exhibited minimal backtracking or revisiting of previously explored areas, indicating high exploration efficiency and effective frontier selection under real-world conditions.

\begin{figure*}[t]
    \centering
    \includegraphics[width=\textwidth]{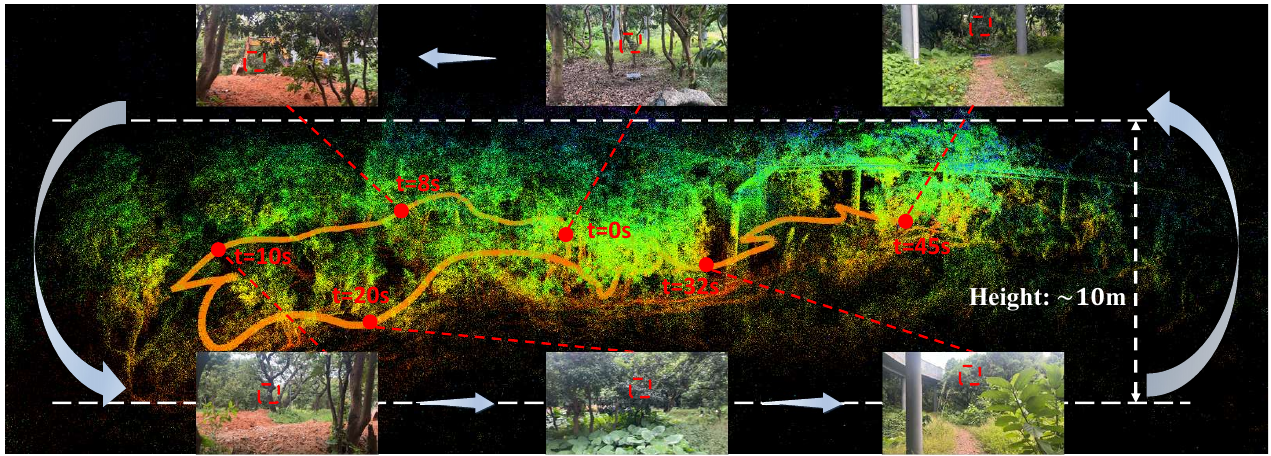}
    \caption{The UAV performs autonomous exploration in a complex and dense forest environment with significant terrain variations, including depressions and uphill slopes.}
    \label{fig: exploration_result_1}
\end{figure*}

Furthermore, we evaluated our method in sloped mountainous regions with significant terrain variations, including depressions and uphill slopes, to assess its robustness and generalization capability. The area measured approximately $50 \times 25 \times 10\ \text{m}^3$. The UAV took off from a depression and conducted a fast exploration of the region. Compared to the previous scenario, this environment was more challenging, featuring denser obstacles such as trees and bushes, which required the UAV to autonomously adapt to the changing terrain and plan trajectories quickly. Fig. \ref{fig: exploration_result_1} shows the executed trajectory along with third-person views of corresponding positions. Throughout the mission, the UAV successfully adapted to the terrain variations and quickly completed the coverage of the environment, without frequent oscillation between altitude layers and while maintaining a high level of exploration consistency. Overall, the UAV completed the exploration in 45 seconds, covering a trajectory of 129.15 meters, further validating the effectiveness of our method.

\section{Conclusion and Future Work}
In this work, we presented a unified mapping framework that maintains spatial consistency between 2D and 3D representations, enabling robotic systems to leverage the advantages of both. Our approach combines the efficiency of 2D maps for fast frontier detection and path planning with the spatial awareness to support terrain-adaptive exploration. This design allows the system to operate effectively in environments with unknown and varying elevations, without requiring prior scale information. To further improve trajectory stability, particularly in uneven outdoor environments, we introduced a terrain-aware cost into the exploration module. This encourages smoother paths and improves energy efficiency during exploration, which is especially beneficial for aerial robots. Extensive simulation and real-world experiments confirmed that our method enables fast and complete coverage in large-scale environments while maintaining high flight speed, demonstrating the framework’s lightweight and efficient characteristics. In future work, we plan to extend this framework to multi-robot systems, enabling scalable and cooperative exploration in more complex scenarios.

\ifCLASSOPTIONcaptionsoff
  \newpage
\fi

\bibliographystyle{IEEEtran}
\bibliography{IEEEabrv,Bibliography}

\vfill

\end{document}